\documentclass[letterpaper]{article} 
\usepackage{aaai2026}  
\usepackage{times}  
\usepackage{helvet}  
\usepackage{courier}  
\usepackage[hyphens]{url}  
\usepackage{graphicx} 
\usepackage{natbib}  
\usepackage{caption} 
\usepackage{algorithm}
\usepackage{algorithmic}

\usepackage{newfloat}
\usepackage{listings}

\usepackage{amsmath} 
\usepackage{multicol}
\usepackage{multirow}
\usepackage{textcomp}
\usepackage[dvipsnames]{xcolor}
\usepackage{subcaption}
\usepackage{amsfonts}
\usepackage{pdfpages}

\DeclareCaptionStyle{ruled}{labelfont=normalfont,labelsep=colon,strut=off} 
\floatstyle{ruled}
\newfloat{listing}{tb}{lst}{}
\floatname{listing}{Listing}
\title{Exploring the Potentials of Spiking Neural Networks for Image Deraining}
\author{
    Shuang Chen\textsuperscript{\rm 1},
    Tomas Krajnik\textsuperscript{\rm 2},
    Farshad Arvin\textsuperscript{\rm 1}, 
    Amir Atapour-Abarghouei\textsuperscript{\rm 1}\thanks{Corresponding Author}, 
}
\affiliations{
    \textsuperscript{\rm 1}Computer Science Department of Durham University, UK\\
    \textsuperscript{\rm 2}Artificial Intelligence Centre of Czech Technical University, Czechia\\

    \{shuang.chen, farshad.arvin, amir.atapour-abarghouei\}@durham.ac.uk\\
    tomas.krajnik@fel.cvut.cz
}

\usepackage{bibentry}

\begin{document}

\maketitle

\begin{abstract}
Biologically plausible and energy-efficient frameworks such as Spiking Neural Networks (SNNs) have not been sufficiently explored in low-level vision tasks. Taking image deraining as an example, this study addresses the representation of the inherent high-pass characteristics of spiking neurons, specifically in image deraining and innovatively proposes the Visual LIF (VLIF) neuron, overcoming the obstacle of lacking spatial contextual understanding present in traditional spiking neurons. To tackle the limitation of frequency-domain saturation inherent in conventional spiking neurons, we leverage the proposed VLIF to introduce the Spiking Decomposition and Enhancement Module and the lightweight Spiking Multi-scale Unit for hierarchical multi-scale representation learning. Extensive experiments across five benchmark deraining datasets demonstrate that our approach significantly outperforms state-of-the-art SNN-based deraining methods, achieving this superior performance with only 13\% of their energy consumption. These findings establish a solid foundation for deploying SNNs in high-performance, energy-efficient low-level vision tasks.
\end{abstract}

\begin{links}
    \link{Code}{https://github.com/ChrisChen1023/VLIF}
\end{links}

\begin{figure}[t]
    \centering
    \includegraphics[width=1\linewidth]{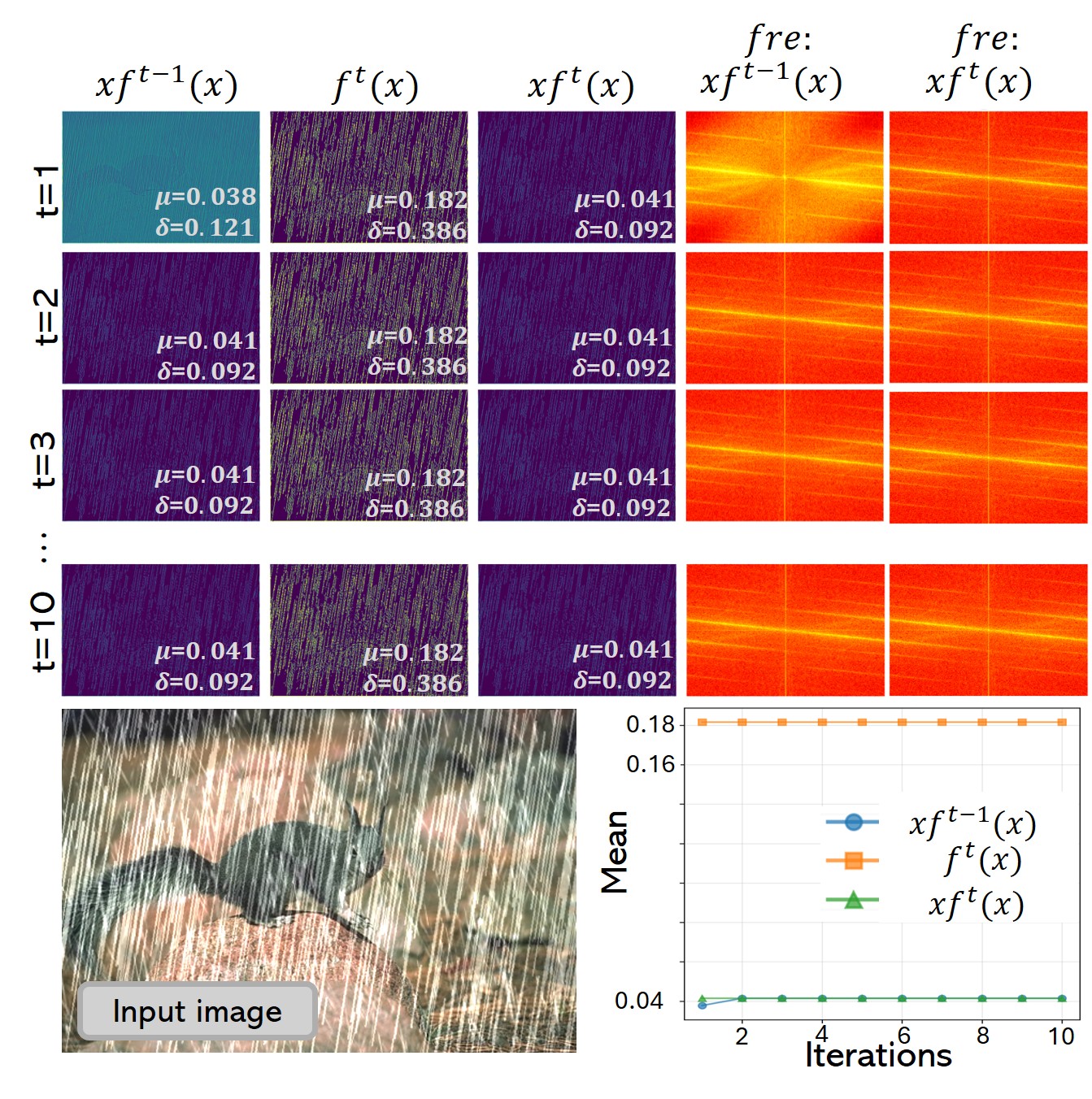}
    \caption{LIF \(f(\cdot)\) highlights high-frequency rain but exhibits frequency saturation with repeated applications after \(t\)=\(1\). 
    \(x\) is embedded features, \(fre\) is frequency spectrums.}
    \label{fig:lif_for_deraining}
\end{figure}

\section{Introduction}

Spiking Neural Networks (SNNs) have recently emerged as highly efficient and biologically plausible alternatives to traditional Artificial Neural Networks (ANNs) and Deep Neural Networks (DNNs) with significant improvements in energy efficiency and computational simplicity. Recent studies have demonstrated the promising potential of SNNs across high-level computer vision tasks, including image classification~\cite{deng2022temporal}, object detection~\cite{luo2024integer} and semantic segmentation~\cite{patel2021spiking}, with effective sequential information processing and representation.

Given the remarkable efficiency and success in high-level visual tasks, a natural question arises: can SNNs also excel in low-level vision tasks? In this paper, we focus specifically on image deraining, a fundamental low-level vision task. Addressing this challenge requires answering a fundamental \textbf{Question} and overcoming a critical \textbf{Obstacle}. The \textbf{Q}uestion is: what distinctive features do SNNs represent in the image deraining context? Unlike convolutional layers, which aggregate local information using weighted filters, the conventional spiking neuron, LIF (Leaky Integrate-and-Fire), integrates discrete spikes over time to trigger neuronal firing events~\cite{maass1997networks}. Such temporal accumulation differs fundamentally from the spatially continuous convolutional operation, potentially leading to distinct feature extraction mechanisms. Although~\cite{song2024learning, su2025bridge, liang2025spikederain, xiao2025spiking} introduce SNNs to vision restoration, the existing works have not addressed this differentiation at the feature representation level, a gap particularly crucial for low-level vision tasks.

To answer this fundamental \textbf{Q}uestion, we analyse and identify two crucial functional properties inherent in LIF neurons: Intensity-Triggered Activation and High-Frequency Indication. We conduct extensive experiments and reveal that the discrete, threshold-triggered spikes of LIF neurons inherently emphasise high-frequency components dominated by high-intensity, high-frequency rain artifacts.

Moreover, our investigation reveals another significant representational bottleneck in conventional LIF neurons: repeated applications of the standard LIF mechanism do not further enrich feature representations beyond an initial high-frequency extraction. Motivated by this insight, we propose the Spiking Decomposition and Enhancement Module (SDEM) and a Spiking Multi-scale Unit (SMU). These novel modules are designed to overcome the saturation effect intrinsic to na\"ive spiking neurons by systematically decomposing input features into complementary high- and low-frequency components. This refines features across multiple spatial scales for a better hierarchical representation of fine-grained rain structures.

The primary \textbf{O}bstacle lies in the inherently pixel-level receptive field of the spiking neuron, which severely limits its capability to capture spatial dependencies necessary for representation-based visual tasks. For instance, in image deraining, accurately identifying and removing rain streaks requires contextual understanding beyond pixel-level perception. Previous attempts to enhance spiking neurons' expressivity through architectural modifications achieve limited success, as they failed to provide spiking neurons with true visual perception capabilities. Other efforts introduced convolutional operations to expand the receptive field in SNNs. Unfortunately, this approach fundamentally contradicts the intrinsic efficiency of SNNs by significantly increasing parameters and computational overhead.

To overcome the identified \textbf{O}bstacle, we draw inspiration from neurophysiological findings showing that neurons in the inferotemporal (IT) cortex of the ventral visual pathway integrate stimuli across spatially extended receptive fields to recognise complex object features~\cite{kobatake1994neuronal}. Motivated by this mechanism, we leverage the cumulative firing behaviour of Leaky Integrate-and-Fire (LIF) neurons to provide visual awareness. However, due to the intrinsic firing-resetting mechanism of LIF neurons, such binary accumulation alone fails to represent rich visual features. To address this, we propose the Visual LIF (VLIF), combining the above insights with Integrate-and-Fire dynamics enhanced by continuous membrane potentials, thus equipping LIF neurons with spatially meaningful receptive fields without introducing additional parameters.

Extensive evaluations across four benchmark deraining datasets confirm the effectiveness and efficiency of our approach. Our proposed method outperforms existing SNN-based models and achieves competitive or superior results compared to state-of-the-art CNN and transformer-based methods, all while significantly reducing computational overhead and energy consumption.

Our primary contributions are summarised as follows:
\begin{itemize}
\item We bridge the gap between SNNs and low-level vision tasks by identifying and characterising the distinctive representational capabilities of Spiking neurons in the context of image deraining.

\item We provide visual perception capabilities to LIF neurons and explicitly elucidate the feature representation mechanism of conventional LIF neurons in image deraining.

\item We propose the novel spiking neuron Visual LIF (VLIF), to incorporate local visual context through spatial aggregation and continuous membrane potentials, overcoming the spatial insensitivity and representational limitations of conventional LIF neurons.

\item Leveraging the VLIF neuron, we introduce a novel image deraining framework comprising the proposed Spiking Decomposition and Enhancement Module (SDEM) and the Spiking Multi-scale Unit (SMU). This framework significantly surpasses the performance of the best existing SNN-based deraining models while consuming only 13\% of their energy.
\end{itemize}

\section{Related Work}
\subsection{Image Deraining}
Single image deraining has rapidly advanced with the application of deep learning, where initial convolutional neural network (CNN)-based methods typically approached deraining as pixel-wise regression tasks~\cite{jiang2020mspfn, zamir2021MPRNet}. More recent methods have employed multi-stage or multi-scale CNN architectures and self-attention mechanisms to improve rain removal performance by capturing intricate patterns and modelling long-range dependencies~\cite{Zamir2022Restormer}. Despite their effectiveness, these CNN and attention-based approaches often incur heavy computational costs, limiting deployment on resource-constrained devices. 
Very recently, a different paradigm was explored with spiking neural networks~\cite{song2024learning}. The SNN-based approach opens a new direction for efficient rain removal on low-power devices.

\subsection{Spiking Neural Network}
Spiking Neural Networks (SNNs) offer energy-efficient computation through sparse, event-driven processing, making them promising alternatives to conventional ANNs. SNNs are typically implemented via ANN-to-SNN conversion or direct training. 
The former maps a trained ANN to an SNN by replacing activations with spiking units~\cite{cao2015spiking_ann, diehl2015fast_ann_relu}, but often requires long simulation time and suffers from performance degradation. In contrast, direct training with surrogate gradients enables end-to-end optimisation and achieves competitive accuracy with much fewer timesteps and lower latency~\cite{neftci2019surrogate}.
In computer vision, SNNs have been applied to high-level tasks such as image classification~\cite{hu2021ms-resnet}, object detection~\cite{luo2024integer} and segmentation~\cite{patel2021spiking}.
However, their use in pixel-level low-level tasks (such as image deraining) remains rare. We explore this under-investigated area and demonstrate that SNNs can be effectively adapted for dense image regression.

\begin{figure*}[ht!]
    \centering
    \includegraphics[width=17.5cm]{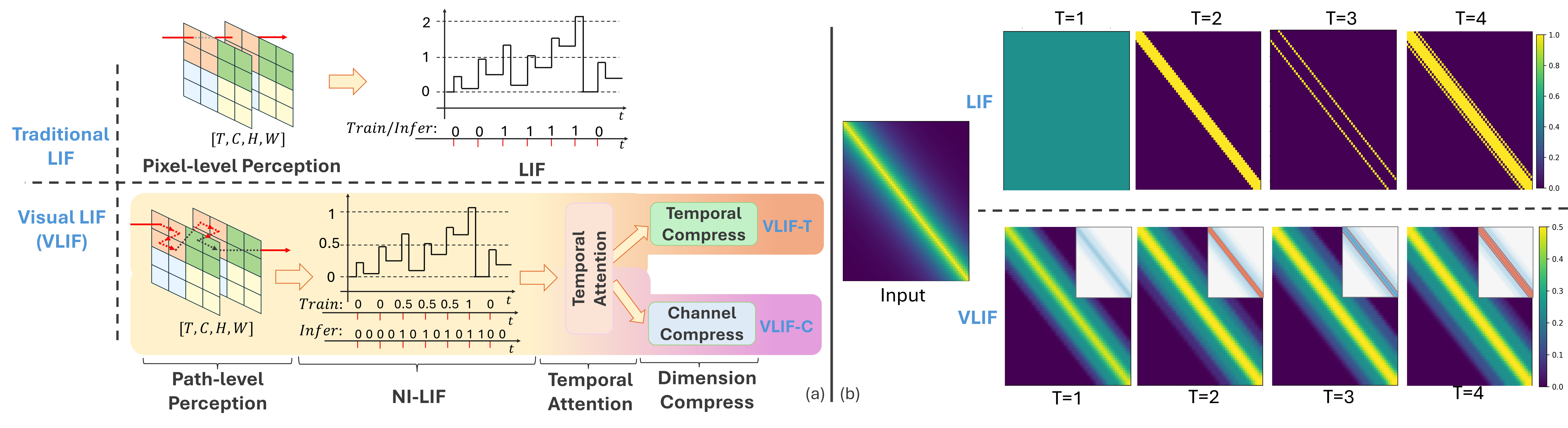}
    \caption{The comparison of LIF and VLIF. (a) shows differences in pipeline, (b) shows differences response areas, in the top-right insets, blue and orange denote the response for LIF and VLIF, respectively.}
    \label{fig:vlif}
\end{figure*}

\begin{figure}[h]
    \centering
    \includegraphics[width=0.95\linewidth]{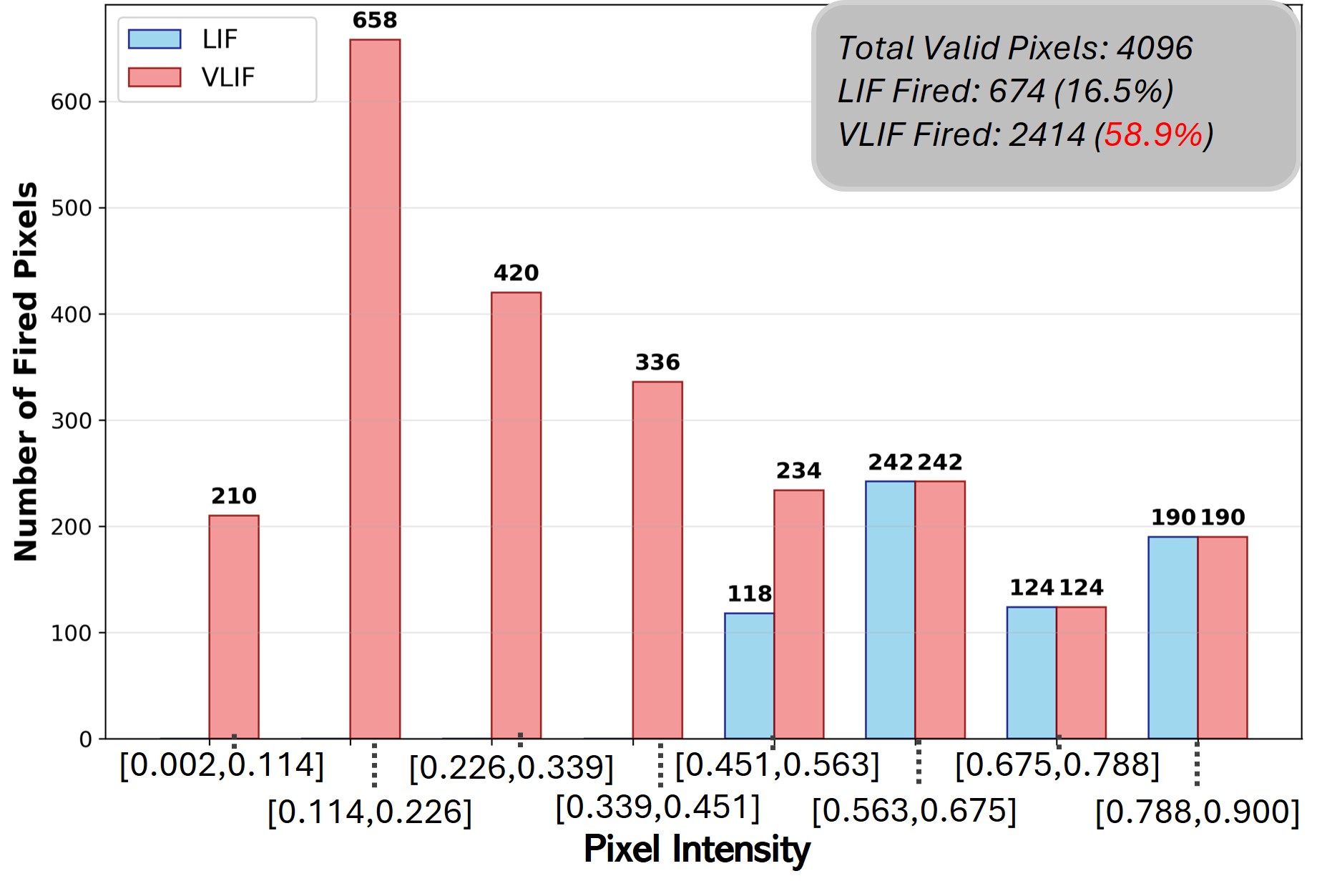}
    \caption{Activation comparison.
    VLIF activates $3.58\times$ more than LIF by incorporating local spatial context, enabling better feature representation in low-response regions.}
    \label{fig:fired_pixels}
\end{figure}

\section{Method}
In image deraining, the key challenge lies in accurately identifying and separating high-frequency rain structures from degraded images.
However, directly applying SNNs to this task suffers from limited visual awareness due to the binary and sequential firing behaviour of standard LIF neurons.

Through empirical analysis, we identify two functional properties of LIF neurons relevant to deraining: a) Intensity-Triggered Activation and b) High-Frequency Indication. Building on this, we derive two fundamental theoretical properties: a) high-frequency selection and b) frequency saturation, which expose a representational bottleneck in standard LIF neurons. These findings motivate our framework, which addresses these limitation through three core components: 1) the Vision-LIF (VLIF) neuron with spatial awareness, (2) the Spiking Decomposition \& Enhancement Module (SDEM) for feature disentanglement and refinement and (3) the Spiking Multi-scale Unit (SMU) for efficient multi-scale representation learning at high-resolution stages.

\begin{figure*}[h]
\begin{center}
\includegraphics[width=16cm]{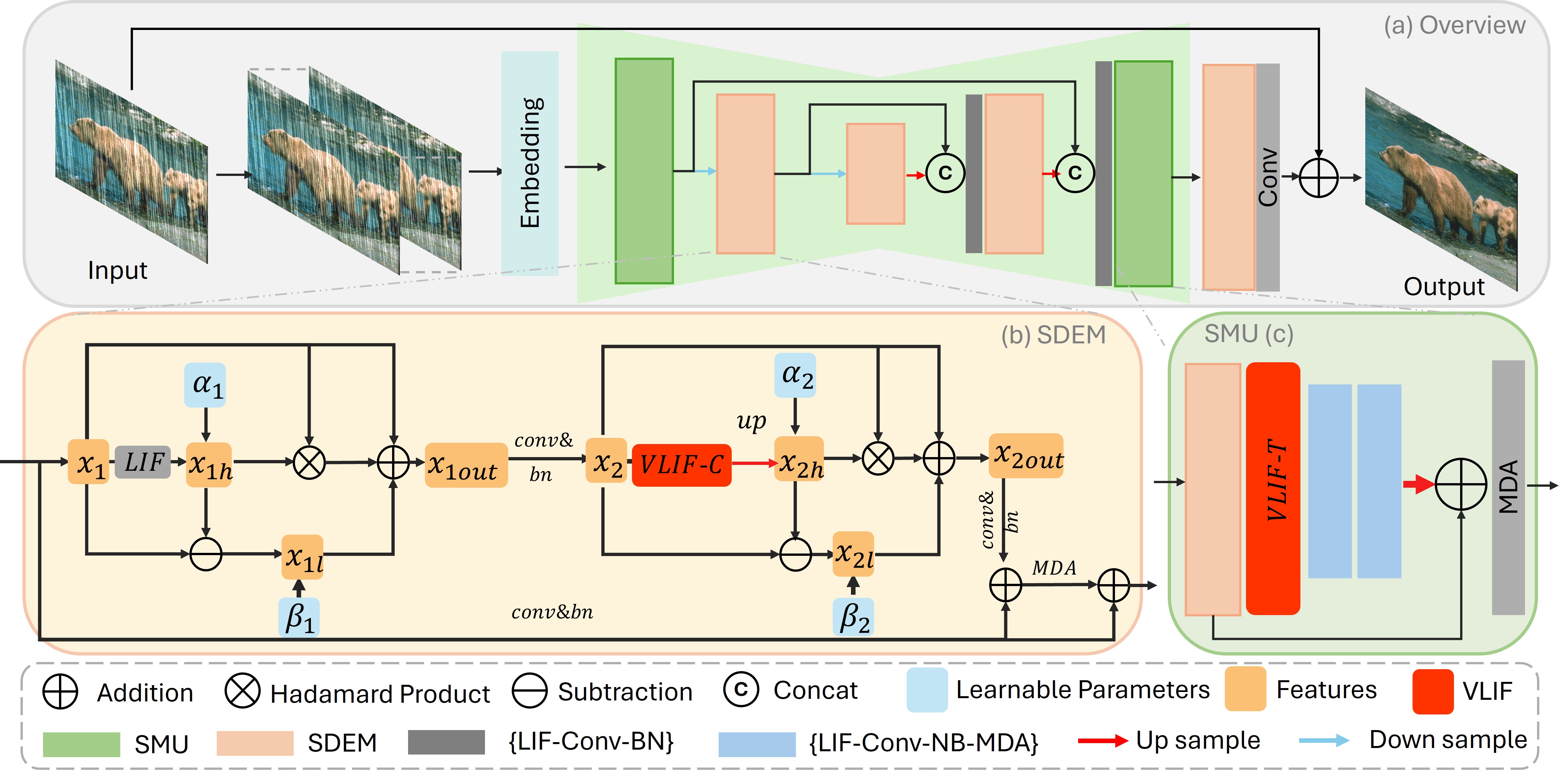}
\end{center}
   \caption{(a) Architecture; (b) Spiking Decomposition \& Enhancement Module (SDEM), (c) Spiking Multi-scale Unit (SMU).}
\label{fig:overview}
\end{figure*}
\subsection{Spiking Neurons for Image Deraining?}
We begin by introducing our key finding: In image deraining, LIF neurons act as high-pass indicators that emphasise high-frequency rain structures. Notably, LIF neurons do not inherently possess high-pass filtering capabilities. Their effectiveness arises from the close alignment between their firing behaviour and the core objective of the deraining task, which is to identify rain structures that typically manifest as high-frequency noise superimposed on the image through residual representation learning.

As the rain streaks dominate this discrepancy between clear and rainy images, 
these regions contribute more significantly to the gradient during backpropagation, leading the network to amplify their intensity in the learned feature representations. The membrane potential dynamics of LIF neurons are inherently sensitive to such localised, high-intensity responses. When the feature intensity at a given spatial location surpasses the neuron's firing threshold $\theta$, a spike is triggered. Consequently, the sparse binary output of the spiking neuron effectively serves as a spatial mask that highlights regions corresponding to high-frequency rain patterns.

Based on this behaviour, we define the spiking neuron as an Intensity-Triggered High-Frequency Indicator in the context of image deraining. This role is characterised by two essential functional properties: 1) \textbf{Intensity-Triggered Activation}: The neuron's output depends solely on whether local feature intensities exceed a predefined threshold. 2) \textbf{High-Frequency Indication}: The neuron's binary outputs mark spatial locations where intensity surpasses the threshold, which typically aligns with high-frequency artefacts such as rain streaks.
We formally define the concept of a high-frequency indicator as follows:
Given a feature representation, $x$, and a LIF neuron $f(\cdot)$ applied $t$ times, denoted as $f^t(x)$, the neuron acts as a high-frequency indicator if it satisfies the following condition:
\begin{equation}
||x \odot f^t(x)||_2 =
\begin{cases}
||HC[x]||_2 = C, & \text{if } t = 1, \\
||x \odot f^{t-1}(x)||_2 = C, & \text{if } t > 1,
\end{cases}
\end{equation}
where $\odot$ denotes the Hadamard product and $HC[x]$ denotes the high-frequency components in the feature space of $x$.

This formulation reveals two critical theoretical properties: 1) \textbf{High-Frequency Highlighting}: At the first application ($t$=$1$), the operator $f(\cdot)$ identifies spatial and temporal positions of high-frequency components through intensity thresholding. The resulting sparse binary mask $f(x)$, when applied to $x$ via element-wise multiplication, enhances the visibility of high-frequency rain artefacts. Validated in Fig.~\ref{fig:lif_for_deraining}, for $t=1$, compared to $fre:x f^{t-1}(x)$, $fre:xf^t(x)$ shows a clear reduction in low-frequency energy. 2) \textbf{Frequency Saturation}: For subsequent applications ($t$$>$$1$), repeated use of $f(\cdot)$ does not produce a more refined or sparser high-frequency representation. The extracted high-frequency content remains equivalent to that obtained at $t=1$. In Fig~\ref{fig:lif_for_deraining}, $\mu$ is fixed after $t$=$1$. This property is mathematically reflected in the invariance for all $t > 1$.


Such properties reveal a fundamental limitation of the standard LIF: repeated applications of the same neuronal mechanism do not enrich the feature representation beyond the initial extraction of high-frequency information. Inspired by this observation, we propose novel modules VLIF, SDEM and SMU, to leverage more diverse feature representations across multiple scales, thereby overcoming the saturation effect and enabling a more effective architecture for image deraining. Next, we introduce the proposed Vision-LIF.
\begin{table*}[!ht]
\centering
\setlength{\tabcolsep}{1mm}
\begin{tabular}{cc|ccccccccccc}
\hline
\multicolumn{2}{c|}{} & \multicolumn{2}{c}{\textbf{Rain12}} & \multicolumn{2}{c}{\textbf{Rain200L}} & \multicolumn{2}{c}{\textbf{Rain200H}} & \multicolumn{2}{c}{\textbf{Rain1200}} & \multirow{1}{*}{\textbf{Params}$\downarrow$} & \multirow{1}{*}{\textbf{FLOPs}$\downarrow$} & \multirow{1}{*}{\textbf{Energy}$\downarrow$} \\
\multicolumn{2}{c|}{\multirow{-2}{*}{\textbf{Methods}}} & {PSNR}$\uparrow$ & {SSIM}$\uparrow$ & {PSNR}$\uparrow$ & {SSIM}$\uparrow$ & {PSNR}$\uparrow$ & {SSIM}$\uparrow$ & {PSNR}$\uparrow$ & {SSIM}$\uparrow$ & \multirow{1}{*}{(M)}  &\multirow{1}{*}{(G)}  &\multirow{1}{*}{($\times10^4$$\mu$J)}  \\ \hline
\multicolumn{1}{c|}{} & PReNet       & 36.32 & 0.9593 & 36.06 & 0.9735 & 27.18 & 0.8698 & 33.39 & 0.9280 & 0.168   & 66.249  & 82.81 \\
\multicolumn{1}{c|}{\multirow{-2}{*}{\textbf{C}}} & Effderain    & 36.12 & 0.9588 & 36.06 & 0.9731 & 26.11 & 0.8341 & 32.85 & 0.9147 & 27.654  & 52.915  & 66.14 \\ \hline
\multicolumn{1}{c|}{} & Restormer    & 37.33 & 0.9672 & 36.62 & 0.9731 & 27.93 & 0.8823 & 31.56 & 0.9009 & 26.10   & 12.33   & 11362 \\
\multicolumn{1}{c|}{\multirow{-2}{*}{\textbf{T}}} & SmartAssign  & 36.87 & 0.9618 & 38.41 & 0.9814 & 27.71 & 0.8536 & 33.11 & 0.9154 & 1.359   & 90.386  & 112.9 \\ \hline
\multicolumn{1}{c|}{} & ESDNet       & 36.38 & 0.9610 & 38.34 & 0.9845 & 28.39 & \textbf{0.8944} & \textbf{32.78} & 0.9126 & 0.165   & 7.320   & 9.150 \\
\multicolumn{1}{c|}{\multirow{-2}{*}{\textbf{S}}} & Ours         & \textbf{36.70} & \textbf{0.9649} & \textbf{38.46} & \textbf{0.9847} & \textbf{28.63} & {0.8935} & 32.74 & \textbf{0.9184} & \textbf{0.088} & \textbf{2.794} & \textbf{1.166} \\ \hline
\end{tabular}
\caption{Quantitative results on four synthetic benchmarks.
\textbf{C}, \textbf{T}, and \textbf{S} represent CNN-based, Transformer-based, and SNN-based methods, respectively.
\textbf{Bold}
denote the best score.
Our method achieves consistently superior performance compared to Restormer, a representative and widely adopted Transformer-based baseline, while substantially reducing computational cost.}
\label{tab1}
\end{table*}

\subsection{Vision-LIF: Spiking Neuron with Visual Awareness}
As discussed above, LIF neurons serve as effective high-frequency indicators, producing sparse binary activations for strong rain artefacts. However, the spatial insensitivity is a drawback. The firing decision at each pixel is made independently and relies solely on the pixel-level intensity. Such a mechanism fails to account for spatially correlated but low-intensity regions (such as smooth peripheries of rain patterns), which are unlikely to be strong enough to trigger the firing threshold. As a result, these semantically meaningful yet low-contrast components are ignored, leading to incomplete structural representations.
To solve this drawback, we introduce \textbf{Vision-LIF (VLIF)}, a modified spiking neuron with local visual awareness. Unlike conventional LIF neurons, VLIF incorporates spatial context into the spike generation process by aggregating feature intensities from neighbourhoods of pixels. The comparison is shown in Fig~\ref{fig:vlif}.

\noindent \textbf{Patch-to-Time Reorganisation}. 
Given the input feature \( X \in \mathbb{R}^{T \times C \times H \times W} \), we first apply a \textit{pixel unshuffle}~\cite{shi2016real} with downsampling factor \( r \), downscale \( X \) to \( X'\in \mathbb{R}^{T \times C\cdot r^2 \times \frac{H}{r} \times \frac{W}{r}}\),
which splits \(X\) into \( r^2 \) non-overlapping spatial patches.
Next, we reshape and permute the tensor by bringing the patch dimension into the temporal axis to have \(X''\in \mathbb{R}^{T \cdot r^2 \times C \times \frac{H}{r} \times \frac{W}{r}}\).
This transformation effectively converts local spatial patches into temporally ordered sequences, which allows the neuron to integrate local visual context over the temporal axis. Therefore, low-intensity pixels within high-response regions are more likely to fire, thus enhancing spatial continuity in rain structure encoding:
\begin{equation}
F(h, w) = \sum_{t=1}^{T \cdot r^2} U_t(h, w),
\end{equation}
where \(F(h,w)\) is the fused output at spatial location (h,w), \(U_t(h,w)\) is the membrane potential at timestep \(t\). This accumulation explicitly incorporates local spatial context into the firing dynamics of VLIF, which will allow even low-intensity pixels, especially those within high-response neighbourhoods, to surpass the firing threshold. In contrast to conventional LIF neurons, which fire strictly based on individual pixel intensity, VLIF facilitates more context-aware activation of semantically relevant but low-contrast features.


\paragraph{Spiking and Stability Challenge}
While VLIF fires based on spatial context, the spatial accumulation carries greater structural variability and finer-grained information. In such cases, binary spikes are insufficient to express continuous intensity differences.
To address this, we consider I-LIF~\cite{luo2024integer}, an integer-valued spiking neuron that generates discrete spike counts for better representation. However, similar to the observation in Spike2Former~\cite{lei2025spike2former}, I-LIF leads to training instability. In our task, this manifests as integer overflow in the output images. We argue that the potential reason is that the large integer values accumulated by I-LIF lead to corrupted or saturated reconstruction results, which fail to represent finer-grained features.

To mitigate this, we adopt virtual timestep normalisation~\cite{lei2025spike2former}, to bound the spike values and stabilise gradient flow, named NI-LIF, by scaling spike magnitudes with $D$ and apply clipping to bound \(x\) to \([x,min, max]\):
\begin{align}
U[t] &= H[t-1] + X[t], \\
S[t] &= \frac{\text{Clip}(\text{round}(U[t]),\ 0,\ D)}{D}, \\
H[t] &= \beta \cdot (U[t] - S[t] \cdot D),
\end{align}
where $U[t]$ is the membrane potential, $S[t]$ is the normalised spike output, $H[t]$ is the updated state, $D$ is the virtual timestep scaling constant, and $\beta$ is the leakage factor.


Following the spiking stage, we apply a temporal attention ~\cite{yao2023attention} to obtain the attended spike features with the shape \( [{r^2}T, C, H/r, W/r] \). We denote this output as \( S_{\text{attn}} \), which is subsequently used for output compression.


\paragraph{Output Compression.}
To compact the additional dimension, we introduce temporal- and channel-based compression to fuse global contextual information.
For \textit{Temporal Compression (VLIF-T)}, we apply a \( r^2 \times 1 \times 1 \) 3D convolution to aggregate information across the temporal dimension:
    \begin{equation}
    S_{\text{TC}} = \text{Conv3D}_{r^2 \times 1 \times 1}(S_{\text{attn}}).
    \end{equation}
    
For \textit{Channel Compression (VLIF-C)}, we flatten the temporal axis and apply a lightweight MLP. This enables flexible channel-wise fusion and allows the model to learn channel correlations across the aggregated temporal features:
    \begin{equation}
    S_{\text{CC}} = \text{MLP}(\text{Flatten}(S_{\text{attn}})).
    \end{equation}

Both variants produce a compressed representation with reduced temporal dimensionality, which is then passed to subsequent network stages.

\subsection{Analysis of more Fired Pixels}
As shown in Fig.~\ref{fig:vlif}b,
we compare VLIF with traditional LIF using a synthetic decay matrix, where diagonal elements have the intensity of 0.9 and off-diagonal elements decay exponentially to simulate the rainy structure.

\cite{ding2025rethinking} finds that the output of the vanilla LIF is poorly informative at the first timestep and only achieves decent performance after integrating subsequent temporal steps, as the LIF with \(t\)=\(1\) suffers from strong membrane potential suppression due to the zero-initialised hidden state. As a result, input features may fail to exceed the firing threshold at \(t\)=\(1\) (Fig.~\ref{fig:vlif}b), even with relatively high intensity. This behaviour limits LIF to represent fine-grained spatial details, which is critical in pixel-level restoration tasks such as image deraining.
In contrast, VLIF normalises spike outputs and encourages early activation, yielding significantly denser responses. As shown in Fig.~\ref{fig:fired_pixels}, LIF achieves only 16.46\% activation, while VLIF reaches 58.94\% activation (\(3.6\times\) improvement) and maintains 25.4\% fire rate even for the lowest input range [0.002, 0.114], where LIF is unable to fire. This showcases VLIF's strength in preserving weak signals and reconstructing fine details through spatio-temporal reorganisation and multi-level quantisation.



\begin{figure*}[t]
    \centering
    \includegraphics[width=0.95\linewidth]{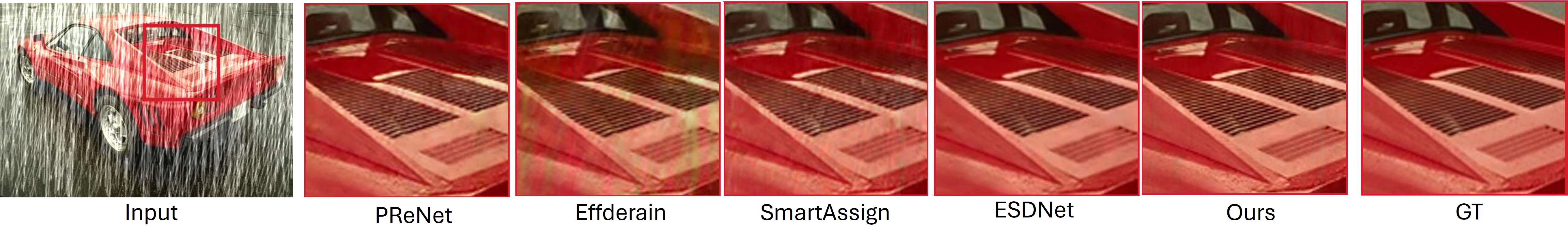}
    \caption{Qualitative comparison on the Rain200H. Our model generates finer-grained details and clearer textures.}
    \label{fig:comp_rain200H}
\end{figure*}
\subsection{Spiking Decomposition \& Enhancement Module}
To further enhance the model's capacity to detect fine-grained rain structures, we propose the Spiking Residual Block (SDEM), which systematically decomposes input features into high- and low-frequency components via spiking neurons. 
%
As shown in Fig.~\ref{fig:overview}(b), SDEM comprises two cascaded stages. Each stage follows an identical processing pipeline but differs in the spiking neuron utilised, LIF for the first stage to detect sharp and strong high-frequency response, while VLIF-C for the second stage for context-aware detection of subtle or blurred rain structures. This progressive design facilitates coarse-to-fine enhancement of rain features. 
Next, we only detail Stage 2 for simplicity.

\begin{table}[!t]
\centering
\small
\setlength{\tabcolsep}{1mm}

\begin{subtable}[t]{1.0\linewidth}
\centering
\small
\begin{tabular}{cccccc}
\hline
Model & Effderain & Restormer&ESDNet & Ours   \\ \hline
PIQE$\downarrow$ & 63.56 & 32.58 & 26.10  &  24.26     \\
BRISQUE$\downarrow$ & 75.28 & 28.55 & 22.75    &  20.99  \\
\hline
\end{tabular}
\caption{Quantitative result for Real-World data.}
\label{tab:rwdata}
\end{subtable}

\begin{subtable}[t]{1.0\linewidth}
\centering
\small
\begin{tabular}{ccccccc}
\hline
Model & T-EMB & SMU & SDEM  & REFINE  & PSNR$\uparrow$   & SSIM$\uparrow$   \\ \hline
(a)    &        &       &        &         & 26.51 & 0.8612 \\
(b)    & $\checkmark$ &       &        &         & 26.59 & 0.8639 \\
(c)    & $\checkmark$ &       & $\checkmark$ &   & 26.96 & 0.8657 \\
(d)    & $\checkmark$ & $\checkmark$ & $\checkmark$ &   & 27.11 & 0.8778 \\ 
Ours   & $\checkmark$ & $\checkmark$ & $\checkmark$ & $\checkmark$ & \textbf{27.21} & \textbf{0.8799} \\
\hline
\end{tabular}
\caption{Ablation study on architectural components.}
\label{tab:abl_each_comp}
\end{subtable}
\begin{subtable}[t]{1.0\linewidth}
\centering
\small
\setlength{\tabcolsep}{1mm}
\begin{tabular}{cccccc}
\hline
Model & LIF & I-LIF & VLIF-C  & VLIF-T  & Ours   \\ \hline
PSNR$\uparrow$  & 21.8362 & 26.19 & 23.10 & 26.19 & 27.21 \\
SSIM$\uparrow$  & 0.6722  & 0.8595 & 0.7719 & 0.8755 & 0.8799 \\
\hline
\end{tabular}
\caption{Ablation study on different LIF variants.}
\label{tab:abl_vlif}
\end{subtable}

\begin{subtable}[t]{1.0\linewidth}
\centering
\small
\begin{tabular}{l|ccccc}
\hline
Metric    & TACL & UIE-WD & SFGNet& UIR-PK & Ours \\
\hline
PSNR$\uparrow$     & 20.99 & 17.80 & 22.68 & \textbf{26.42} & 25.69\\
SSIM$\uparrow$     & 0.782 & 0.760 & 0.585  & 0.866 & \textbf{0.8817}\\
Params (M)$\downarrow$  & 28.29 & 14.46 & 1.30  & 1.84 & \textbf{0.088}\\
FLOPs (G)$\downarrow$  & 240.06 & 102.76 & 163.16  & 27.34 & \textbf{2.7938}\\
\hline
\end{tabular}
\caption{Quantitative comparison on EUVP.}
\label{tab:comp_euvp_filtered}
\end{subtable}

\caption{(a) Quantitative real-world results, (b) ablation study, (c) impact of LIF variants, (d)  UIE task performance.}
\label{tab:combined_ablation}
\end{table}

\paragraph{Stage-wise Feature Decomposition.}
Let the input to the second stage be denoted as $X \in \mathbb{R}^{T \times C \times H \times W}$. We begin by applying the VLIF-C on \(X\), after employing a bilinear interpolation for upsampling, we get a high-frequency indicator \(X_h\) scaled by a learnable parameter $\alpha$, then subtract \(X_h\) to obtain the complementary low-frequency representation \(X_l\):
\begin{equation}
X_h = \alpha\cdot\text{VLIF-C}(X), \quad X_l = X - X_h,
\end{equation}
where $X_h$ serves as a structural high-pass mask and $X_l$ retains the low-frequency component. Additionally, we use $X_h$ to extract the high-frequency feature $X_h'$. A learnable scalar $\beta_2$ is adopted to scale the low-frequency path:
\begin{equation}
X_h'=X_h\cdot X, \quad X_l' = \beta_2 X_l.
\end{equation}

We then aggregate all components with a residual connection, followed by a convolution and batch normalisation~\cite{zheng2021tdbn} to refine the features:
\begin{equation}
X_{\text{refined}} = \text{BN}(\text{Conv}(X_{h}' + X_{l}' + X)).
\end{equation}

\paragraph{Residual and Attention Fusion.}
We include a residual shortcut path to further preserve the input feature distribution and ensure gradient stability:
\begin{align}
    X_{\text{shortcut}} &= \text{BN}(\text{Conv}(X)), \\
    X_{\text{refined}}' &=X_{\text{refined}} + X_{\text{shortcut}}. 
\end{align}

Additionally, a multi-dimensional attention module (MDA)~\cite{yao2023attention} enhances spatio-temporal and channel interactions. Then the final SDEM output is given:
\begin{align}
X_{\text{MDA}} &= \text{MDA}(X_{\text{refined}}),\\
X_{\text{SDEM}} &= X_{\text{refined}'} + X_{\text{shortcut}} + X_{\text{MDA}}.
\end{align}


\subsection{Spiking Multi-scale Unit (SMU)}
Inspired by the fact that multi-scale modelling at the shallow level can enhance representational capacity~\cite{cui2023image}, we introduce a lightweight multi-scale module, \textbf{SMU}, which leverages the proposed SDEM and a VLIF-T, to incorporate fine-grained rain cues from both local and extended neighbourhoods with minimal computational overhead, offering an efficient multi-scale enhancement mechanism in highest-resolution stages of both encoder and decoder.

As illustrated in Fig.~\ref{fig:overview}(c), SMU first receives input feature \( x \in \mathbb{R}^{T \times C \times H \times W} \). \(x\) then is processed by the proposed SDEM, followed by a VLIF-T neuron that further downsamples the spatial resolution to $[H/r, W/r]$ via patch-to-time reorganisation.
The downsampled feature is then passed through two stacked Spiking Units (SU), each composed of a LIF spiking neuron, batch normalisation, convolution, and a multi-dimensional attention (MDA) module~\cite{yao2023attention}. Two SUs further extract high-frequency representations at lower spatial resolutions. The features are subsequently upsampled back to $[T, C, H, W]$ using bilinear interpolation.
To retain shallow spatial information, a residual connection bypasses intermediate layers, with merged features refined by an MDA block before the next step.

\section{Experiments}
This section presents datasets, metrics, our main experimental results on draining, ablation studies, and a generalisation study on another low-level vision task, Underwater Image Enhancement (UIE). The implementation details, training setting, and hyperparameter tuning (e.g.\ time step, number of blocks) are provided in supplementary materials. 




\subsection{Datasets and Metrics}
To comprehensively evaluate the performance, we conduct experiments on five widely used synthetic rain datasets (Rain12~\cite{li2016GMM}, Rain200L~\cite{yang2017JORDER}, Rain200H~\cite{yang2017JORDER}, Rain1200~\cite{zhang2018diddata}) and the Real-World rain dataset~\cite{ren2019PReNet}.
All datasets are used following the original train/test splits (shown in the supplementary material with training iterations). We evaluate model performance for synthetic data on two widely used metrics, Peak Signal to Noise Ratio (PSNR)~\cite{hore2010PSNR} and Structural Similarity (SSIM)~\cite{wang2004SSIM}, shown in Tab.~\ref{tab1} (reproduced from~\cite{song2024learning}).
We retrain ESDNet based on their official setting, as their pre-trained models are inaccessible. Perception-based Image Quality Evaluator (PIQE)~\cite{venkatanath2015PIQE} and Blind/Referenceless Image Spatial QUality Evaluator BRISQUE~\cite{mittal2012no} are used for the real-world data, shown in Tab.~\ref{tab:rwdata}.
Following~\cite{hu2021spikingresnet, song2024learning}, energy is calculated with: 12.5pJ per FLOP, 77fJ per SOP (synaptic operation), and 3.7pJ per spike-based Sign operation (calculated using energy per spike). All ablation studies are conducted on RainH200 for 30K iterations.

\subsection{Experiment Results}
In Tab.~\ref{tab1}, our method fully exploits the potential of spiking neurons for image deraining. Compared with the latest SNN-based method ESDNet, our approach achieves superior performance while using only \textbf{53\%} of its parameters (0.088M vs. 0.165M), \textbf{38\%} of its FLOPs (2.79G vs. 7.32G), and \textbf{13\%} of its energy consumption ($1.17 \times 10^4\,\mu$J vs. $9.15 \times 10^4\,\mu$J). Moreover, our model further narrows the performance gap between SNNs and strong ANN-based baselines. Qualitative results in Fig.~\ref{fig:comp_rain200H} show that our method generates finer-grained and more structurally coherent outputs compared to ANN models such as Effderain~\cite{guo2021efficientderain} and SmartAssign~\cite{wang2023smartassign}, while requiring only \textbf{0.3\% / 6.5\%} of their parameter counts (0.088M vs. 27.65M / 1.36M) and \textbf{1.8\% / 1.0\%} of their energy consumption ($1.17 \times 10^4\,\mu$J vs. $6.61 \times 10^5\,\mu$J / $1.13 \times 10^6\,\mu$J). Lower PIQE and BRISQUE from Tab.~\ref{tab:rwdata} and comparison from Fig.~\ref{fig:realworld-rain} underscore our robustness in real-world scenarios.


\subsection{Ablation Study}

{\flushleft{\textbf{Effectiveness of Each Components}.}}
As shown in Tab.~\ref{tab:abl_each_comp}, model (a) is the baseline, where the spiking neuron degrades to a vanilla LIF within a standard U-Net structure. Adding Temporal EMBedding (b) brings slight improvement.
Model (c) further introduces SDEM, which results in a significant PSNR improvement. This shows effectiveness to extract and enhance fine-grained rain features. Comparing (c) and (d), the inclusion of the SMU notably improves SSIM, indicating better preservation of spatial structure. Finally, the refinement block composed of four stacked SDEM modules provides further gains to achieve the best overall performance.

\begin{figure}[t]
    \centering
    \begin{subfigure}[t]{1.0\linewidth}
        \centering
        \includegraphics[width=\linewidth]{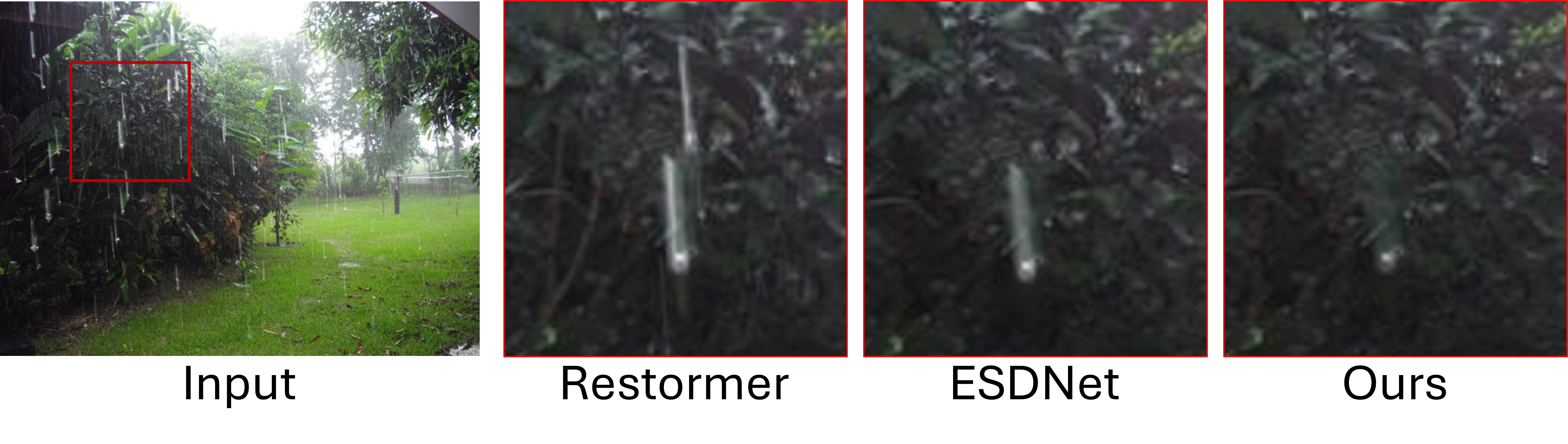}
        \caption{Real-world rain image.}
        \label{fig:realworld-rain}
    \end{subfigure}
    \hfill
    \begin{subfigure}[t]{1,0\linewidth}
        \centering
        \includegraphics[width=\linewidth]{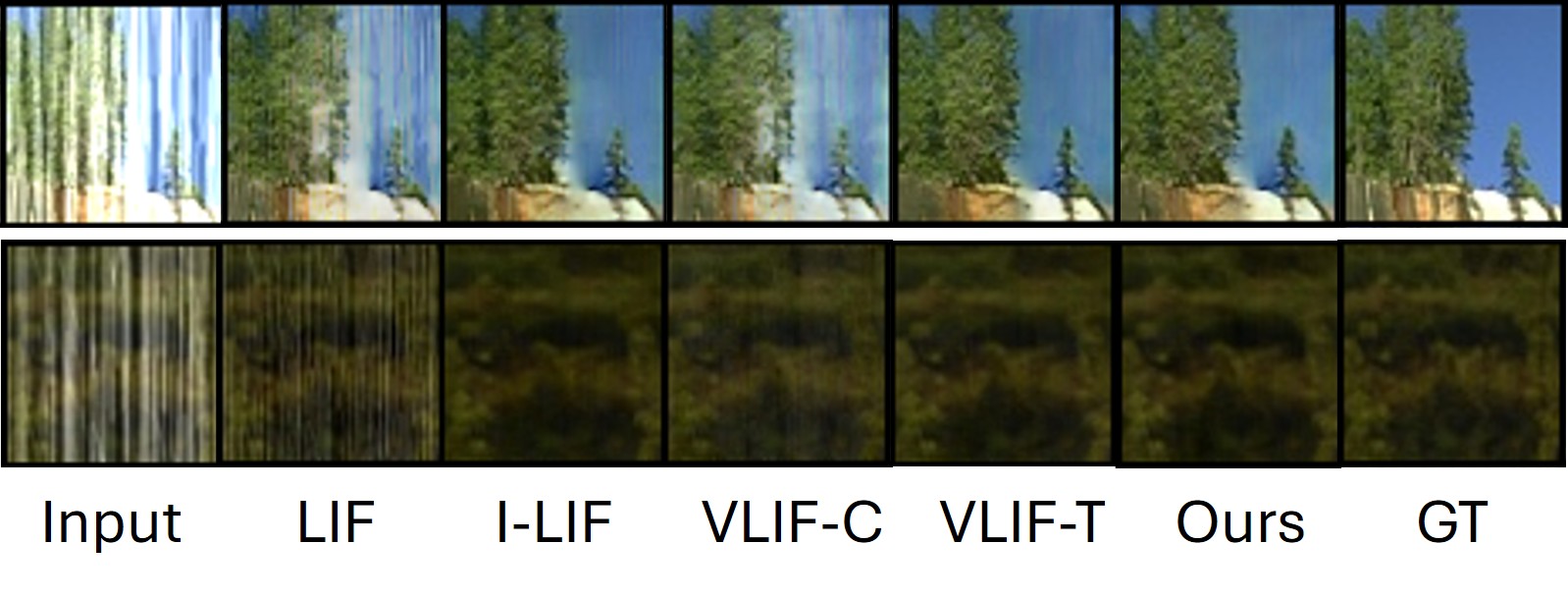}
        \caption{Spiking neuron variant results.}
        \label{fig:lifs}
    \end{subfigure}

    \hfill
    \begin{subfigure}[t]{1.0\linewidth}
        \centering
        \includegraphics[width=1\linewidth]{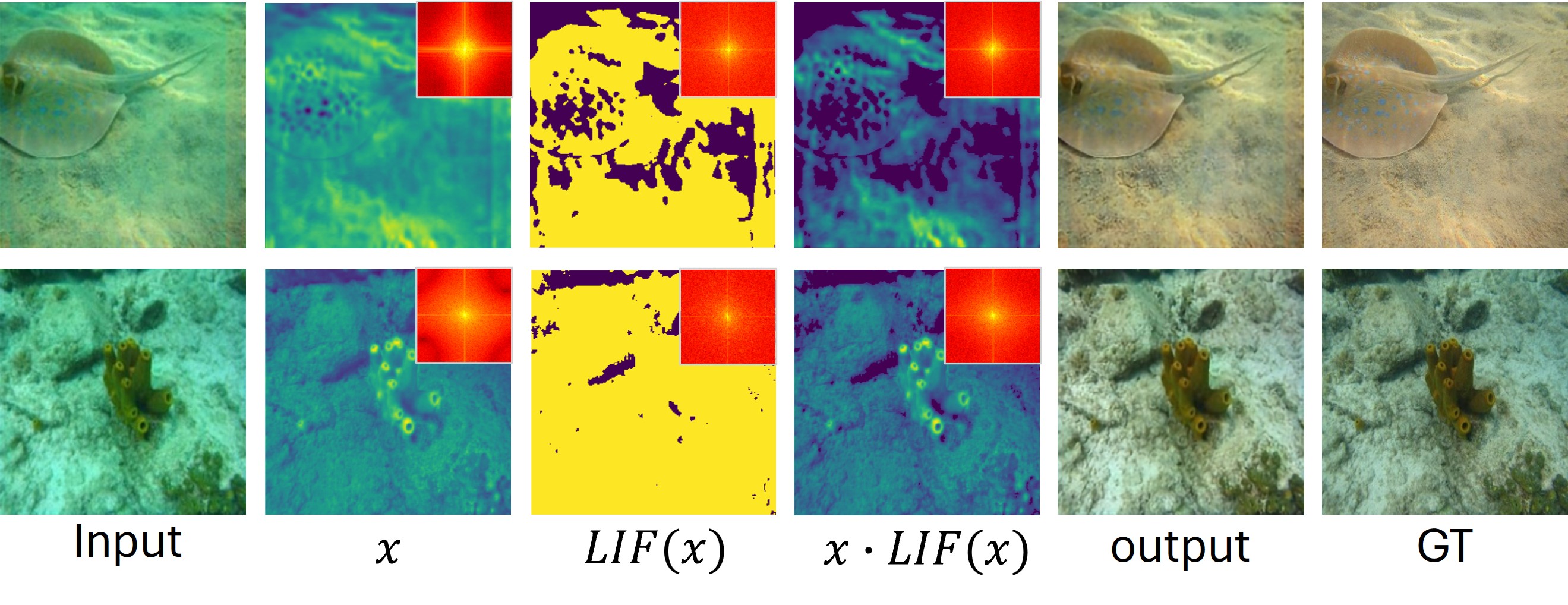}
        \caption{Visualisation of UIE task. Left to right: input image, feature before LIF, binary spike map, LIF-highlighted feature, our result, and ground truth. Top right corners show the frequency spectrums.}
        \label{fig:uie}
    \end{subfigure}
    \caption{(a) Real-world rainy image; (b) Spiking neuron variants; (c) Visualisation on the UIE task.}
    \label{fig:real_lif_comparison}
    
\end{figure}
{\flushleft{\textbf{Effectiveness of VLIF}.}}
We evaluate different spiking variants by replacing the VLIF neurons in our full model with LIF, I-LIF, VLIF-C, and VLIF-T. Table~\ref{tab:abl_vlif} shows that, stacking standard LIF leads to significant degradation (19.7\%$\downarrow$ in PSNR and 23.6\%$\downarrow$ in SSIM) compared to the full model. This is because repeated use of basic LIF provides limited features, as it suffers from \textit{frequency saturation}. The drop observed with VLIF-C indicates that channel-only compression is insufficient for handling extended sequences. While I-LIF, VLIF-T, and our hybrid design achieve comparable quantitative scores, qualitative results in Fig.~\ref{fig:lifs} show that our full model produces cleaner and more detailed outputs, benefiting from joint temporal and channel compression.

\subsection{Generalisability}

To further validate the task-oriented behaviour and generalisability of spiking neurons in image restoration, we train our model on the EUVP~\cite{islam2020fast} Underwater Image Enhancement (UIE) dataset, where the goal is to correct large-scale low-frequency degradation from light scattering and colour shifts. As shown in Tab.~\ref{tab:comp_euvp_filtered}, our method surpasses recent UIE approaches~\cite{liu2022twin, ma2022wavelet, zhao2024toward} while using the least computation. Consistent with our frequency-domain analysis (Fig.~\ref{fig:uie}), LIF neurons act as low-frequency indicators in UIE, highlighting the task-adaptive nature of spiking neurons in image restoration.

\section{Conclusion}
This work demonstrates the potential of Spiking Neural Networks for single-image deraining. We answer the key \textbf{Q}uestion of how spiking neurons represent rain artefacts, showing that in deraining they behave as task-driven, intensity-triggered high-frequency indicators. To overcome the \textbf{O}bstacle of spatial insensitivity in conventional LIF neurons, we propose Visual LIF (VLIF) to enable spatial awareness. In addition, our Spiking Decomposition \& Enhancement Module and Spiking Multi-scale Unit provide hierarchical, multi-scale representations of rain patterns.

Experimentally, our method outperforms existing SNN-based approaches and achieves performance comparable to ESDNet while using only 13\% of its energy. Nonetheless, a performance gap between SNNs and ANNs remains, largely due to differences in representational richness and optimisation maturity. Closing this gap is an important direction for future work.


\bigskip

\bibliography{aaai2026}


\section*{Supplementary material}

# Supplementary Material - Exploring the Potentials of Spiking Neural Networks for Image Deraining

## Architecture

We adopt a classic encoder-decoder architecture motivated by its effectiveness in hierarchical representation learning. The network is composed of a Temporal Embedding Head, an encoder-decoder body, which involves two SMUs (Spiking Multi-scale Unit) at the shallowest levels for multi-scale learning with less complexity, and a Refinement Block built from SDEMs (Spiking Decomposition & Enhancement Modules). Based on the experiments, we skipped the refinement block on RainH200 and Rain1200 to achieve better performance. Skip connections and residual learning are employed to enhance detail preservation.

Given an input image $[H, W, 3]$ with $H \times W$ spatial dimensions and RGB channels, we perform direct input encoding for the static image to obtain a sequence of input tokens $X = \{x_t\}_{t=1}^T$. This sequence is fed into a Temporal Embedding Block, which uses a $3 \times 3$ convolution followed by temporal-wise attention (Yao et al. 2023) to regulate temporal dependencies and enrich feature representation. The embedded features are then passed into the encoder-decoder body, hierarchically models spatial representation through the proposed SMU and SDEMs. A final refinement block composed of four SDEMs further enhances feature quality. The output is then compressed and passed through a $3 \times 3$ convolution layer to generate a residual image, which is added back to the degraded input to produce the final restored image.

## Implementation Details

Unless otherwise specified, all comparison experiments are conducted under consistent training settings. Our deraining architecture adopts a three-level encoder–decoder structure. The proposed **Spiking Multi-scale Unit (SMU)** is placed at the shallowest level (Level 1), while the deeper levels (Levels 2 and 3) employ stacked **Spiking Decomposition Modules (SDEMs)** with a stage-wise configuration of $[4, 10, 2]$. An additional refinement stage contains 4 SDEMs to further enhance reconstruction quality. For the Vision-LIF (VLIF) spiking neuron, we use a patch-to-time downsampling factor of 2, and set the virtual timestep scaling constant $D = 4$. The training patch size is set to $64 \times 64$, and the batch size is 12. We follow (Song et al. 2024) by using the Sigmoid function as a surrogate gradient for backpropagation training and adopting SSIM as the reconstruction loss for model optimisation. All experiments are conducted using the PyTorch framework on a single NVIDIA A6000 GPU.

To facilitate the training of spiking neural networks (SNNs) with non-differentiable binary activations, we adopt a surrogate gradient approach. Specifically, the Sigmoid function is used as a gradient proxy during backpropagation, allowing effective gradient flow through discrete spiking neurons. The surrogate function and its gradient are defined as:

$$\sigma(x) = \frac{1}{1 + e^{-\alpha x}}, \tag{1}$$

$$\sigma'(x) = \alpha \cdot \sigma(x) \cdot (1 - \sigma(x)), \tag{2}$$

where $\alpha$ is a hyper-parameter controlling the steepness of the surrogate gradient.

For model optimization, we employ the Structural Similarity Index (SSIM) loss to encourage the preservation of fine structural details in derained outputs. The loss is defined as:

$$\mathcal{L}_{ssim} = 1 - \text{SSIM}(\mathcal{S}(\mathcal{X}), \mathcal{Y}), \tag{3}$$

where $\mathcal{S}(\cdot)$ denotes the output of our network, $\mathcal{X}$ is the input rainy image, and $\mathcal{Y}$ is the ground-truth clean image.

| Datasets | Rain12 | Rain200L | Rain200H | Rain1200 | Real-World |
|---|---|---|---|---|---|
| train | 0 | 1800 | 1800 | 12K | 0 |
| test | 12 | 200 | 200 | 1200 | 34 |
| iter | 0 | 300K | 300K | 200K | 0 |

Table 1: Training and test splits for each dataset, and the corresponding training iterations. The model trained on Rain200L and Rain200H to test Rain12 and Real-World Rain, respectively.

## Datasets

We conduct experiments on five widely used synthetic rain datasets (Rain12 (Li et al. 2016), Rain200L (Yang et al. 2017), Rain200H (Yang et al. 2017), Rain1200 (Zhang and Patel 2018)) and the Real-World rain dataset (Ren et al. 2019). We use the model trained on Rain200L to test Rain12, and the model trained on Rain200H to test Real-world rain. For the UIE (underwater image enhancement) task, we train our model on EUVP (Islam, Xia, and Sattar 2020) dataset for 400 epochs.

| | Methods | Rain12 | | Rain200L | | Rain200H | | Rain1200 | | Average | | Params↓ | FLOPs↓ | Energy↓ |
|---|---|---|---|---|---|---|---|---|---|---|---|---|---|---|
| | | PSNR↑ | SSIM↑ | PSNR↑ | SSIM↑ | PSNR↑ | SSIM↑ | PSNR↑ | SSIM↑ | PSNR↑ | SSIM↑ | (M) | (G) | (uj) |
| C | RESCAN | 35.48 | 0.9499 | 33.82 | 0.9547 | 26.22 | 0.8219 | 32.60 | 0.9271 | 32.03 | 0.9134 | 0.149 | 32.119 | $4.014 \times 10^5$ |
| | PReNet | 36.32 | 0.9593 | 36.06 | 0.9735 | 27.18 | 0.8698 | 33.39 | 0.9280 | 33.24 | 0.9327 | 0.168 | 66.249 | $8.281 \times 10^5$ |
| | RCDNet | 37.67 | 0.9612 | 38.84 | 0.9854 | 28.98 | 0.8889 | 32.68 | 0.9189 | 34.54 | 0.9386 | 2.958 | 194.501 | $2.431 \times 10^6$ |
| | MPRNet | 37.55 | 0.9665 | 39.82 | 0.9863 | 29.94 | 0.8999 | 34.50 | 0.9369 | 35.46 | 0.9474 | 3.637 | 548.651 | $6.858 \times 10^6$ |
| | Effderain | 36.12 | 0.9588 | 36.06 | 0.9731 | 26.11 | 0.8341 | 32.85 | 0.9147 | 32.78 | 0.9202 | 27.654 | 52.915 | $6.614 \times 10^5$ |
| | NAFNet | 37.53 | 0.9639 | 39.48 | 0.9847 | 29.19 | 0.8880 | 34.69 | 0.9367 | 35.22 | 0.9433 | 29.102 | 16.064 | $2.008 \times 10^5$ |
| | SFNet | 36.11 | 0.9503 | 39.50 | 0.9850 | 29.75 | 0.9008 | 34.51 | 0.9383 | 34.97 | 0.9437 | 13.234 | 124.439 | $1.555 \times 10^6$ |
| T | DRT | 37.74 | 0.9674 | 38.81 | 0.9830 | 28.67 | 0.8796 | 33.88 | 0.9283 | 34.78 | 0.9395 | 1.176 | 166.045 | $2.075 \times 10^6$ |
| | ELFformer | 35.06 | 0.9420 | 38.85 | 0.9800 | 28.93 | 0.8852 | 33.54 | 0.9360 | 34.09 | 0.9358 | 1.221 | 23.667 | $2.958 \times 10^5$ |
| | HPCNet | 36.84 | 0.9639 | 39.14 | 0.9847 | 29.17 | 0.8962 | 34.46 | 0.9378 | 34.90 | 0.9378 | 1.411 | 29.540 | $3.692 \times 10^5$ |
| | Restormer | 37.33 | 0.9672 | 36.62 | 0.9731 | 27.93 | 0.8823 | 31.56 | 0.9009 | 33.36 | 0.9309 | 26.10 | 12.33 | $113.62 \times 10^6$ |
| | SmartAssign | 36.87 | 0.9618 | 38.41 | 0.9814 | 27.71 | 0.8536 | 33.11 | 0.9154 | 34.03 | 0.9281 | 1.359 | 90.386 | $1.129 \times 10^6$ |
| S | ESDNet | 36.38 | 0.9610 | 38.34 | 0.9845 | 28.39 | 0.8944 | **32.78** | 0.9126 | 33.97 | 0.9283 | 0.165 | 7.320 | $9.150 \times 10^4$ |
| | Ours | **36.70** | **0.9649** | **38.46** | **0.9847** | **28.63** | **0.8935** | 32.74 | **0.9184** | **34.13** | **0.9404** | **0.088** | **2.7938** | $\mathbf{1.166 \times 10^4}$ |

Table 2: Quantitative results on four synthetic benchmarks. **C**, **T**, and **S** represent CNN-based, Transformer-based, and SNN-based methods, respectively. **Bold** denote the best score. Our method achieves consistently superior performance compared to Restormer, a representative and widely adopted Transformer-based baseline, while substantially reducing computational cost.

| $T \times D$ | FLOPs (G) | Energy ($\mu$J) | PSNR (dB) |
|---|---|---|---|
| $(1 \times 1)$ | 0.77 | $9.38 \times 10^3$ | 25.06 |
| $(1 \times 2)$ | 0.77 | $5.75 \times 10^3$ | 25.88 |
| $(1 \times 4)$ | 0.77 | $4.51 \times 10^3$ | 26.67 |
| $(2 \times 1)$ | 1.45 | $1.66 \times 10^4$ | 25.20 |
| $(2 \times 2)$ | 1.45 | $9.38 \times 10^3$ | 26.03 |
| $(2 \times 4)$ | 1.45 | $6.89 \times 10^3$ | 26.77 |
| $(4 \times 1)$ | 2.79 | $3.12 \times 10^4$ | 25.21 |
| $(4 \times 2)$ | 2.79 | $1.66 \times 10^4$ | 26.12 |
| $(4 \times 4)$ **Ours** | 2.79 | $1.16 \times 10^4$ | 27.21 |
| $(6 \times 1)$ | 4.14 | $4.57 \times 10^4$ | 25.34 |
| $(6 \times 2)$ | 4.14 | $2.39 \times 10^4$ | 26.17 |
| $(6 \times 4)$ | 4.14 | $1.64 \times 10^4$ | 27.40 |

Table 3: Impact of different time steps (T) and quantization steps (D) on Rain200H dataset.

## Hyperparamrter Tuning

Based on the setting of ablation stuies, as shown in Table 3, increasing the temporal step ($T$) generally improves image restoration quality, indicating that a longer temporal range helps the model extract more comprehensive spatiotemporal features. Similarly, increasing the quantization steps ($D$) also leads to a notable enhancement in output quality, likely due to finer representation of pixel-level variations. However, both higher $T$ and $D$ lead to increased computational cost and energy consumption. To maintain a good balance between performance and efficiency, we choose $T = 4, D = 4$ as the default setting in our experiments.

## More Comparison

Table 2 showcases more comprehensive comparison. Our proposed SNN-based method achieves competitive performance across all datasets, narrowing the performance gap with state-of-the-art ANN-based approaches (CNN-based and Transformer-based methods). This trend is consistent with our conclusion: similar to high-level vision tasks (Luo et al. 2024), spiking neural networks demonstrate the potential to approach or even match the performance of their ANN counterparts, while offering substantial improvements in computational efficiency and energy consumption. Nevertheless, a noticeable gap still remains in terms of absolute reconstruction accuracy, especially when compared to the strongest Transformer-based models such as MPRNet and NAFNet. Bridging this residual performance gap remains an important direction for future research in SNN-based low-level vision tasks.


\end{document}